# Improved Hopping Control on Slopes for Small Robots Using Spring Mass Modeling

Heston Roberts
Department of Electrical and Computer Engineering
Georgia Southern University
Statesboro, GA, USA
hr07051@georgiasouthern.edu

Pronoy Sarker
Department of Electrical and Computer Engineering
Georgia Southern University
Statesboro, GA, USA
ps12160@georgiasouthern.edu

Sm Ashikul Islam
Department of Electrical and Computer Engineering
Georgia Southern University
Statesboro, GA, USA
si02348@georgiasouthern.edu

Min Gyu Kim
Department of Electrical and Computer Engineering
Georgia Southern University
Statesboro, GA, USA
mkim@georgiasouthern.edu

*Abstract*— Hopping robots often lose balance on slopes because the tilted ground creates unwanted rotation at landing. This work analyzes that effect using a simple spring–mass model and identifies how slope-induced impulses destabilize the robot. To address this, we introduce two straightforward fixes: adjusting the body's touchdown angle based on the slope and applying a small corrective torque before takeoff. Together, these steps effectively cancel the unwanted rotation caused by inclined terrain, allowing the robot to land smoothly and maintain stable hopping even on steep slopes. Moreover, the proposed method remains simple enough to implement on low-cost robotic platforms without requiring complex sensing or computation. By combining this analytical model with minimal control actions, this approach provides a practical path toward reliable hopping on uneven terrain. The results from simulation confirm that even small slope-aware adjustments can dramatically improve landing stability, making the technique suitable for future autonomous field robots that must navigate natural environments such as hills, rubble, and irregular outdoor landscapes.

## I. Introduction

Hopping robots usually jump and land on flat ground. When the ground is inclined or sloped, landing becomes harder. The main reason is that slope changes how the robot's leg pushes and how it hits the ground. This can create unwanted rotation such that the robot's body can tip over or start spinning which might cause the robot to fall. In the paper by Yim et al. (2020), they showed that controlling angular momentum or how fast the body spins during takeoff and landing helps robots jump precisely [1]. Their robot could jump on flat surfaces and control its rotation in mid-air. In this report, we extend their idea to sloped or inclined surfaces.

Our goal is to investigate and improve the dynamics of hopping robots on inclined surfaces by focusing on leaning and flight control mechanisms. We introduce a new mathematical model and control rule that makes landing stable even when the ground is not flat.

Previous research has extensively investigated the mechanisms that enable robots to obtain these precisions. Early studies demonstrated that underactuated systems such as acrobat can achieve sliding and hopping gaits using energy-based control methods [2] [3]. Here, precise regulation of stance phase torques determined both flight trajectory and landing success. Work in [4] introduced a deadbeat control framework for running robots which provides a biologically inspired feedback design that allows for rapid stabilization after perturbation. This indicated that stance phase timing directly influences recovery dynamics of robots. Further extension of this framework is presented in [5] which formulates a three-dimensional spring mass model that incorporates time-based deadbeat control laws. This formation regulates stance time and ground reaction forces to achieve robust steering and running even in uncertain environments.

Another curvature bounded analysis shows motion plans remain dynamically feasible under turning constraints [6]. The constraints provide critical insights into safe path planning for the robot. The stance phase balance can be maintained through real time feedback of contact forces during jump initiation and recovery [7]. Energy based storage mechanisms have also been refined to introduce magnetic spring-based actuation systems [8]. These are capable of efficient energy recycling during stance compression, enabling higher jumping frequencies without increasing actuator demand. Mechanical energy shaping during stance is the key to maintaining stability and to minimizing control effort. A complementary approach shows that open loop parallel elastic actuation can improve energy efficiency in running monopods [9]. Finally, precise co-contraction control of antagonistic muscle groups can generate stable force patterns which can mirror stance phase dynamics in robotic systems [10].

Research on hopping robots spans a wide range of modeling and control strategies. Foundational simplified hopper models established the basic stance–flight dynamics and the principles of energy exchange [17]. Later work introduced more advanced actuation and structural designs, including elastic mechanisms that use passive vibration to increase lift efficiency [18]. Other studies examined stability through feedback-based hopping control for single-leg platforms [15], as well as adaptive height regulation methods that enable agile and repeatable jumps under varying conditions. More recent efforts have demonstrated high-performance hopping using improved actuators, trajectory shaping, and real-time feedback to achieve robust takeoff and landing behaviors [16]. Comparative evaluations of one-dimensional hopper designs have also shown important trade-offs between mechanical architecture and control strategy, influencing stability and efficiency across different operating conditions.

## II. Background And Motivation Behind the Work

Early studies demonstrated that underactuated robots can achieve hopping through a combination of energy-based methods and torque-driven control strategies. These works showed that the timing of the stance phase and proper regulation of joint torques play critical roles in ensuring stable takeoff and controlled landing. Building on this foundation, researchers introduced deadbeat and spring–mass models that enabled robust running, accurate steering, and fast recovery from external disturbances. Additional advancements incorporated real-time contact-force feedback, which improved both jump initiation and post-impact stabilization.

Bio-inspired actuation systems and elastic or magnetic mechanisms further enhanced energy efficiency, allowing robots to reach greater jump heights with minimal actuator effort. However, despite substantial progress, most precision-leaping research has focused primarily on flat ground. As a result, the dynamics of landing on sloped or inclined surfaces remain insufficiently explored, highlighting a gap that motivates the present study.

## III. Theory Behind Small Robot Hopping

The simplest hopping robot has two moving parts: a body with mass $m_b$ and a foot with mass $m_f$, whose centers of mass are connected by a spring with stiffness $k$. Each part may be treated as its own system in the force analysis. When the robot is airborne, the only forces acting on each part are the weight and the spring force. However, when the robot is grounded, there exists a normal force acting on the foot perpendicular to the surface. In ideal conditions, static friction will counteract the horizontal component of the normal force on inclined surfaces.

Because the forces are dependent on the spring's elongation and whether the robot is airborne, the acceleration is not constant. This means in a continuous-time environment, calculus is required to find the velocity and position of each part. However, because our simulation will use discrete time, we may approximate the kinematics via a constant acceleration system with initial time $t_0$ and final time $t_0 + \Delta t$. Because the sampling time $t$ will be small, the change in acceleration will be negligible.

Therefore, for every sample we may update the velocity $\vec{v}[n]$ and position $\vec{r}[n]$ of the centers of mass of each part with these kinetic equations,

$$\vec{v}[n] = \vec{v}[n-1] + \vec{a}[n-1]\Delta t \quad \vec{r}[n] \\ = \vec{r}[n-1] + \vec{v}[n-1]\Delta t + \frac{1}{2}\vec{a}[n-1](\Delta t)^2 \quad (1)$$

where $n$ is the sample number and $\vec{a}[n]$ is the acceleration. Before proceeding with the next sample, we must update the acceleration via force analysis. Generally, according to Newton's second law of motion, the acceleration of an object is the net force divided by its mass. Thus, when the robot is midair the acceleration of the body and foot are respectively

$$\vec{a_b}[n] = -g\hat{z} - \frac{k}{m_b}\vec{s}[n] \quad \vec{a_f}[n] = -g\hat{z} + \frac{k}{m_f}\vec{s}[n] \quad (2)$$

where $\vec{s}[n]$ is the elongation at sample $n$. Notice how this is a vector, as the spring force also depends on the direction in which the spring faces. This is also not necessarily the difference between $\vec{r}_b$ and $\vec{r}_f$, as we must also account for the equilibrium position.

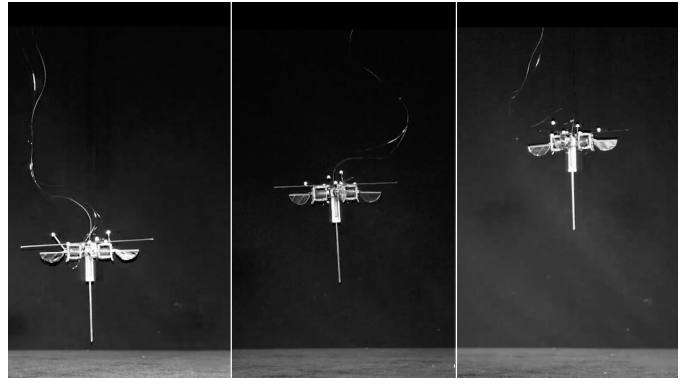

Fig. 1. Hopping robot, adapted from [https://news.mit.edu/2025/hopping-gives-tiny-robot-leg-up-0409]

However, we may define $\vec{d}_n$ as the difference in positions if we then add the spring force and if the spring is fully compressed,

$$\vec{a_b}[n] = -g\hat{z} - \frac{k}{m_b}\big(\hat{d}[n] - d_0\hat{d}[n]\big)\vec{a_f}[n] \\ = -g\hat{z} + \frac{k}{m_f}\big(\hat{d}[n] - d_0\hat{d}[n]\big) \quad (3)$$

where $d_0$ is the equilibrium length and $\hat{d}[n]$ is the normal vector of $\vec{d}[n]$.

When the robot hits the surface, the normal force will ideally eliminate the momentum of the foot. The body's momentum will not change because it has no direct interaction with said surface. This is an inelastic collision because spring absorbs the robot's kinetic energy. In a continuous-time system, this impact involves a net force following a Gaussian curve as a function of time. However, in this discrete-time simulation we can describe the average net force as the change in momentum divided by the sampling time. The velocity at the end of the interval ($\vec{v_f}[n]$) is zero, and the mass cancels when converting from force to acceleration. Therefore, the acceleration of the foot at impact is

$$\vec{a_f}[n] = -\frac{\vec{v_f}[n-1]}{\Delta t} \quad (4)$$

which checks out when substituting into equation 1 because both sides would equal zero.

From there, the normal force and the frictional forces balance other forces acting on the foot, so the foot remains stationary until the spring returns to equilibrium. At this point, all the spring's energy will be converted back to kinetic energy, launching the robot back into the air.

In spring mass configuration, the slope angle directly changes how these forces project along the surface. Even though the spring still acts along the leg axis, the ground reaction no longer aligns with the robot's nominal vertical direction, creating an imbalance in horizontal and vertical impulses. As a result, the foot experiences a small but significant tangential component at impact, which can introduce rotation in the body even if it was initially aligned. This mechanism becomes more pronounced at higher slopes or higher landing speeds, making it essential to account for the altered force geometry when analyzing stability on inclined terrain.

## IV. MATHEMATICAL FORMULATION OF SPRING MASS MODEL

### A. Spring Mass Model's Parameters

Few things happen during a hop. First, the robot pushes off (stance phase). Next it flies in the air (flight phase), and finally it lands (impact phase). In flight, there are no external torques, so angular momentum is conserved. However, the robot's spin rate doesn't change. When landing on a slope, the ground applies a quick impulse at the foot causing a sudden change in angular momentum. We represent this as

$$\Delta L_{impact} = r_\perp J_n \quad (5)$$

where $r_\perp$ is the lever arm between the foot and the center of mass, and $J_n$ is the ground impulse. Therefore, we need to define the touchdown angle that results in zero net torque. If, $\alpha_{real}$ is how much robot leans relative to the slope, $\alpha_d$ is the desired angle when the robot touches the slope and $\varphi$ is the slope angle (positive uphill). Then, the desired landing body angle is

$$\alpha_d = \alpha_{real} + \varphi \quad (6)$$

and in flight

$$\alpha(t_{td}) = \alpha_0 + \omega_0 T \quad (7)$$

where $\alpha_0$ is takeoff body angle, $\omega_0$ is rotation rate set at takeoff, and $T$ is the flight time. To land with the correct angle, the robot needs to satisfy equation (8):

$$\omega_0 = \frac{\alpha_d - \alpha_0}{T} \quad (8)$$

This means if we know how long the robot will be in the air ($T$) as well as what angle we want robot to land at ($\alpha_d$), we can then calculate the rotation rate ($\omega_0$) we need at takeoff. At landing, the sloped ground pushes the foot in a tilted direction. This adds an impact angular impulse:

$$\Delta L_{impact} = r_\perp(\varphi) J_n(\varphi) \quad (9)$$

$$J_n = m(1 + e) v_{n,td} \quad (10)$$

Equation (10) is approximation for $J_n$, where $e$ is restitution or bounciness and $v_{n,td}$ is velocity of the foot toward the ground just before impact. So, a steeper slope or faster landing speed increases the impulse, resulting in more unwanted tilt.

### B. Proposed Two-Step Slope Compensation Method

We fix this with two slope aware control steps. First by choosing the right body angle before landing, and second by applying a small correction torque before takeoff.

$$\alpha_d^* = \left[\alpha_d \beta(\varphi) - A(\varphi) + \frac{1}{T}\alpha_0\right] / \left[\beta(\varphi) + \frac{1}{T}\right] \quad (11)$$

Here, $A(\varphi)$ is the base impact torque from the slope, $B(\varphi)$ is how much the torque changes if angle is changed, and $1/T$ is robot's resistance to rotate during flight. This equation gives a simple rule to pick the best lean angle before landing depending on slope and robot properties.

Even after setting the right angle, a small leftover rotation may still occur. We cancel it by adding a small angular impulse just before takeoff. This is like pre-loading the robot's rotation to cancel what the slope will cause later.

$$\Delta L_{pre} = -L^- + \Delta L_{impact} \quad (12)$$

To apply for it,

$$\tau = \frac{\Delta L_{pre}}{t_s} \quad (13)$$

where $t_s$ is a short time, like 0.02 s, at the end of the stance phase. This small torque ensures the robot lands with almost no rotation, and it can be applied by a tail motor or reaction wheel.

## V. SIMULATION RESULTS AND DISCUSSION

In the MATLAB simulation, the robot begins its motion with the hemispherical foot positioned 60 cm above the inclined surface, whereas the body is placed an additional 10 cm above the foot. The system uses a lightweight structure, with the body mass set to 95 g and the foot mass kept to just 5 g, featuring a small 5 mm radius to capture realistic contact behavior. The leg is modeled as a spring with a stiffness of 500 N/m and an equilibrium (rest) length of 10 cm, allowing it to store and release energy during each hop. To accurately represent interaction with the 30° slope, the simulation includes both static and kinetic friction coefficients, assigned values of 0.6 and 0.42 respectively. The robot's motion is simulated over a 3-second window with a high sampling rate of 200 kHz, ensuring precise tracking of rapid changes in velocity, spring compression, and impact forces throughout the hopping cycle.

Three simulation stages are presented to evaluate hopping behavior on a slope. First, the path without phase control shows vertical landing with no corrective strategy, leading to increasing horizontal drift caused by slope-induced rotation. Second, the path after applying step 1 demonstrates the effect of using only the slope-aware touchdown angle adjustment, which reduces drift but leaves some residual motion. Finally, the path after applying both steps combines touchdown angle correction with a small pre-takeoff torque, effectively canceling unwanted rotation and enabling stable, in-place hopping on the inclined surface. These three stages collectively illustrate the progressive improvements achieved by the control approach.

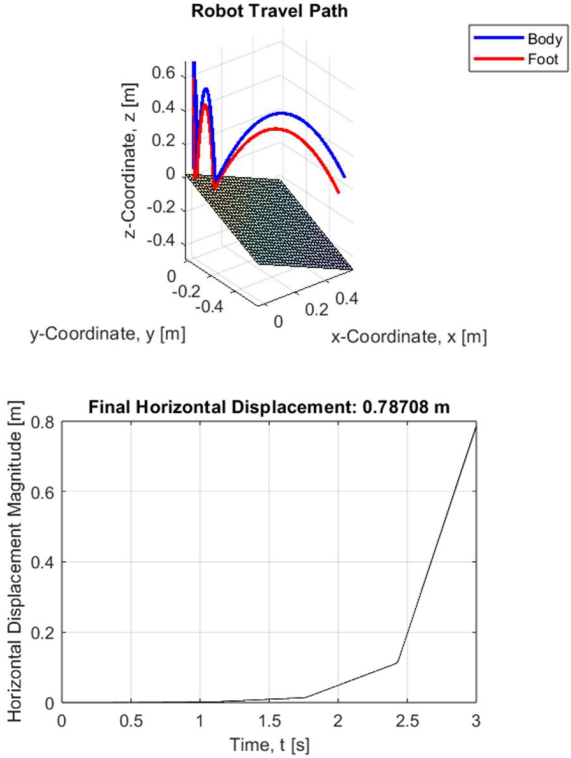

Fig. 2. Simulation results before applying either step.

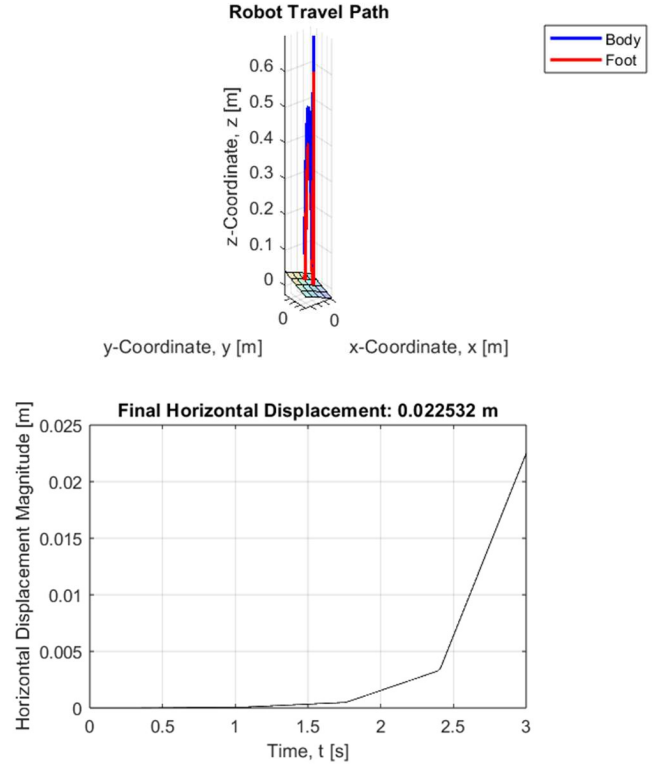

Fig. 3. Simulation results after applying step 1, but before step 2.

### A. Path Without Phase Control

Figure 2 shows the behavior of the hopping robot if it lands perfectly vertically and does not experience any corrective torque. By observing that the time periods between each bounce are nearly the same, we can conclude that the angular impulse increases the horizontal velocity at each bounce. This jeopardizes the robot's ability to continue hopping, let alone traverse its surroundings as intended.

### B. Path After Applying Step 1

The plot shows that applying only the touchdown angle correction significantly reduces horizontal drift, keeping the robot's body and foot paths much closer to vertical. Although some residual motion remains; the hopping becomes noticeably more controlled compared to the no–phase-control case.

By having the robot land with a tilt of 0.006° against the slope, the system already experiences noticeably improved results as shown in Figure 3. Unlike in the first simulation where the robot drifts about 79 cm horizontally, the adjusted landing angle reduces that drift to only 2.25 cm over the same duration, representing a 97% decrease. This demonstrates that even a very small angle change can significantly counteract the slope-induced rotational impulse.

However, slope-aware angle selection alone cannot fully eliminate the remaining sideways motion because a robot cannot land with perfect precision every cycle. Small errors in angle, timing, or impact force still accumulate and create gradual drift. Therefore, additional corrective action becomes necessary. Applying a small torque while the robot is grounded provides an active way to cancel the residual rotation and stabilize the hopping motion, even when the touchdown angle is not exact.

### C. Path After Applying Both Steps

*Fig. 6.* shows that combining both the touchdown angle correction and the pre-takeoff torque eliminates horizontal drift, allowing the robot's body and foot to hop almost perfectly in place on the slope. The trajectory remains nearly vertical for every bounce, demonstrating full cancellation of slope-induced rotation and stable repetitive hopping.

In this simulation, we had the robot start with the same angle as before. However, this time an additional force reacts to any horizontal velocity $\overrightarrow{v_{horiz}}$ by working against it. This force is described as

$$\vec{F} = -\gamma \overrightarrow{v_{horiz}} \qquad (14)$$

with $\gamma$ being the reactivity of the compensator. We chose a reactivity of 7 kg/s and thus achieved little to no horizontal movement, as displayed in Figure 4. After three seconds the robot moves 13 μm horizontally, which is 99.9% less than if only step 1 is applied. This proves that despite the uncertainty in choosing the perfect touchdown angle, it is possible for the robot to bounce in place on this slope using this corrective force.

Though, step 1 is still useful because if the robot lands with a bad angle, the compensator must use higher reactivity to balance the unwanted torque. Not only does this use more power, but it can also lead to oscillation and/or instability. However, if the robot consistently lands with good angles, then the compensator may correct the robot's movement with little reactivity, thus making it more efficient and reliable.

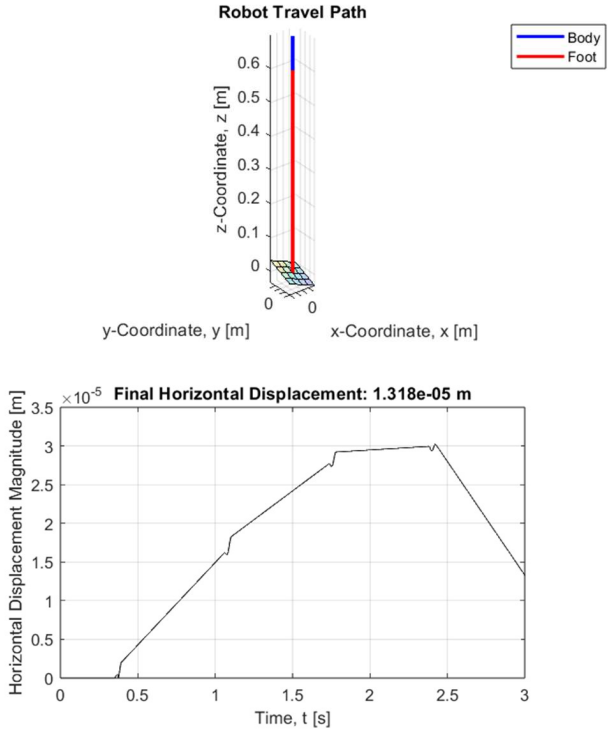

Fig. 4. Simulation results after applying both steps.

While these simulation results are promising, transitioning from an ideal 2D model to a physical 3D environment introduces several practical challenges. Specifically, the current simulation relies on an instantaneous impact model. However, real world collisions involve complex material deformation and varying contact durations which can affect torque application. Furthermore, moving to 3D terrain will require the system to manage not only pitch but also roll variations to prevent lateral tipping. Physical factors such as sensor noise, actuator latency and non-ideal surface friction must be accounted for in hardware deployments to maintain the precision observed in these numerical experiments.

## VI. Improvements Upon Prior Works

The comparison between the previous work [1] and this improved method highlights several key advancements in slope hopping control. Earlier approaches were designed primarily for flat terrain and therefore could not operate reliably on inclined surfaces. As a result, they were unable to cancel the slope-induced rotational impulse that destabilizes the robot during landing. The improved method overcomes these limitations by introducing two essential components: a slope-aware touchdown angle adjustment and a corrective torque applied at takeoff. Together, these additions enable the robot to counteract the unwanted rotation generated by sloped ground contact. Therefore, the improved approach maintains near-zero horizontal drift and supports stability.

TABLE I
COMPARISON BETWEEN PRIOR WORK AND THE CURRENT STUDY

| Topics | Previous Work [1] | SMM |
|---|---|---|
| Works on slopes | No | Yes |
| Cancels slope-induced rotation | No | Yes |
| Touchdown angle adjustment | No | Yes |
| Corrective torque at takeoff | No | Yes |
| Maintains near-zero horizontal drift | No | Yes |

## VII. Conclusion

By adjusting the body angle before landing and applying a small corrective torque at takeoff, the robot effectively cancels the unwanted rotation generated by sloped ground contact, enabling smooth, stable hopping and precise landings on inclined surfaces. The proposed method is mathematically well-founded, physically intuitive, and practical for real robotic systems due to its simplicity and low computational demand. The results demonstrate that even minimal adjustments during the stance and flight phases can dramatically improve stability on non-level terrain. Future work can further strengthen this approach by validating the model across a wider range of simulations, integrating it into hardware experiments, and extending the framework to full 3D terrain involving both pitch and roll variations.

Additionally, incorporating adaptive control strategies would allow the robot to respond intelligently to unknown or dynamically changing slopes. Overall, this work establishes a strong foundation for robust slope-aware hopping control and paves the way for more capable and versatile legged robots.

Furthermore, future research may focus on overcoming the practical limitations of current precision leaping frameworks in unstructured outdoor environments. A particularly promising direction is the integration of reinforcement learning (RL) to develop adaptive torque control strategies. By training a model on diverse sets of incline data, an RL-based compensator could automatically refine the robot's landing angle and pre-takeoff torque in real-time. This will significantly reduce rotational instability without the need for pre-defined mathematical models for every unique terrain.